\documentclass{llncs}
\usepackage[utf8]{inputenc}
\usepackage{graphicx}
\usepackage[export]{adjustbox}
\usepackage{amssymb}
\usepackage{bm} 
\usepackage{amsmath}
\usepackage{booktabs}
\graphicspath{{./images/}}
\begin{document}

\title{Predicting Novel Views Using \protect\\ Generative Adversarial Query Network}
\author{Phong Nguyen-Ha\inst{1} \and 
Lam Huynh\inst{1} \and
Esa Rahtu\inst{2} \and
Janne Heikkila\inst{1}}
\authorrunning{Phong et al.}

\institute{Center for Machine Vision and Signal Analysis, University of Oulu, Finland \and
Tampere University, Finland \\
\email{phong.nguyen@oulu.fi}}
\maketitle

\begin{abstract}
    The problem of predicting a novel view of the scene using an arbitrary number of observations is a challenging problem for computers as well as for humans. This paper introduces the Generative Adversarial Query Network (GAQN), a general learning framework for novel view synthesis that combines  Generative Query Network (GQN) and  Generative Adversarial Networks (GANs). The conventional GQN encodes input views into a latent representation that is used to generate a new view through a recurrent variational decoder. The proposed GAQN builds on this work by adding two novel aspects: First, we extend the current GQN architecture with an adversarial loss function for improving the visual quality and convergence speed. Second, we introduce a feature-matching loss function for stabilizing the training procedure. The experiments demonstrate that GAQN is able to produce high-quality results and faster convergence compared to the conventional approach.  
\end{abstract}

\section{Introduction} \label{Introduction}
Humans are easily able to build a mental understanding of the 3D scene based on just 2D images. With such ability, we can effortlessly imagine unseen views of 3D scenes and objects which is currently extremely challenging for computer based systems. Instead of reconstructing the scene as an explicit 3D model, humans can approximate new views by combining images obtained from nearby poses. Such task of predicting an image from a novel view point, given a limited set of other images from the same scene, is referred as a novel view synthesis in computer vision literature. 

Novel view synthesis can be considered as a fundamental problem in computer vision and it is still being studied actively by the community. Despite of the tremendous progress obtained during the years, the problem is still far from being solved. There are several reasons making novel view synthesis extremely challenging. First, a perfect solution would require knowledge of the full light field of all visible objects which is usually not possible to obtain due to occlusions and limited number of samples. Second, due to
\begin{figure}[ht]
    \centering
    \includegraphics[width=\textwidth]{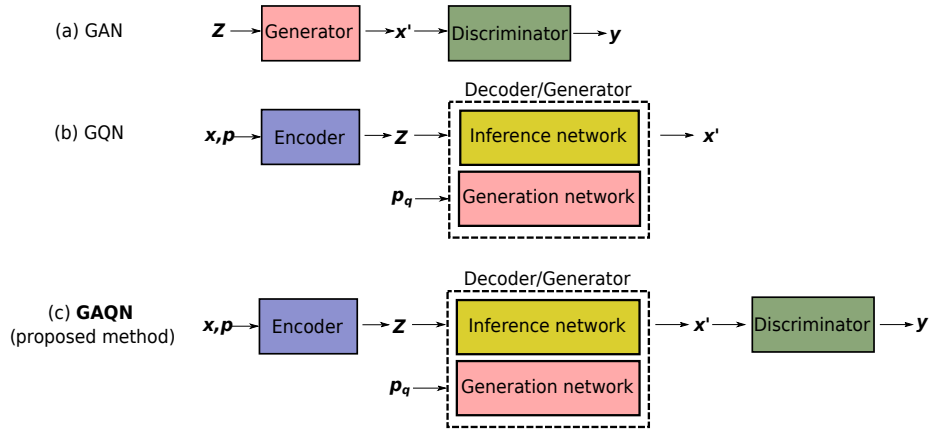}
    \caption{Illustration of the structure of GAN \cite{Goodfellow:2014:GAN:2969033.2969125}, GQN \cite{DeepMind} and our proposed \textbf{GAQN} where $x$, $p$ and $p_q$ are input images, camera poses and queried poses, respectively. $z$ is the latent representation, $y$ is labelled output as real/fake and finally, $x$' is the generated image.}
    \label{fig1}
\end{figure}
short distance to the scene geometry, even  small changes in the viewpoint may lead to substantial changes in the image appearance. Finally, large parts of the scenes are texture-less and textured areas are typically small. Accurate rendering of novel views can be a useful component for many machine vision applications such as robotics and augmented reality. \par
The outline of this paper is as follows. In Section~\ref{RelatedWorks}, we present works related to novel view synthesis methods. A brief background of two previous works that inspired our method is presented in Section~\ref{Background}. In Section~\ref{GAQN_withLSGANloss} and Section~\ref{GAQN_FeatureMatching}, we introduce our Generative Adversarial Query Network with feature matching loss. In Section~\ref{Experiments}, we show our experimental results and discuss the effectiveness of adversarial training and feature matching loss to the previously proposed methods. Finally, we conclude our research in Section~\ref{Conclusion}. 

\section{Related work} \label{RelatedWorks}
In this paper, we propose a solution to the task of novel view synthesis when multiple source images are given. 
Early works on this domain manage to cope with small viewing angle changes by interpolation \cite{intro_interpolation}, warping \cite{intro_interpolation2} or stereo reconstruction \cite{intro_interpolation3}. When the input cameras are further separated from each others, these methods do not perform well due to the sparsity of sampled plenoptic function. Traditional structure-from-motion, structure-from-depth and multi-view geometry methods \cite{SFM_2,Unsupervised_Learning_of_3D_Structure,Learning_Probabilistic_Latent_Space,Online_Structure_Analysis} predict novel views through the estimated 3D structure (point clouds, mesh clouds or a collection of predefined primitives) of the environment. Although, these methods show good results on abundant source data, they are unable to recover the desired target views with a limited number of input images due to the ambiguity of 3D environments. Moreover, estimating the full 3D scene structure may be more challenging problem than that of synthesizing novel images from new viewpoints. 

Recent works on deep generative model such as Variational Auto Encoder (VAE) \cite{VAE}, Generative Adversarial Network (GAN) \cite{Goodfellow:2014:GAN:2969033.2969125} and their variants \cite{WGAN,karras2018progressive,LS_GAN_new,SAGAN} have demonstrated remarkable results in generating highly photo-realistic novel images. Based on these results, one could expect that similar architectures would be applicable to predict 3D structure of the environment. However, the results demonstrated so far, are far from the desired quality. For instance, the viewpoint transformation networks explicitly learn the relationship between input views of the same 3D scene, but the result is limited in scale such as predicting novel views of an individual rotated objects \cite{Viewpoint_network_3,FastViewSynthesiswithDeepStereoVision,Multi-view2NovelviewECCV2018,Viewpoint_network_2,Viewpoint_network_1,Viewpoint_network_4} or predicting a small displacement between stereo camera views \cite{Displacement_2,Displacement_1}. 

A recent generative model that has shown promises in learning representation for 3D scenes structure is Generative Query Network (GQN) \cite{DeepMind}. GQN makes use of an iterative latent variable density model \cite{convolutional_DRAW} to generate images of the 3D scene. Using multiple source images as input, GQN presents an end-to-end learning framework that generates the novel view from the queried pose by leveraging learned knowledge of the 3D scene representation. However, GQN is known for large memory consumption and the predicted novel views are sometimes blurry.

\section{Background}\label{Background}
In this section, we provide the reader with a brief background of Generative Adversarial Networks (GANs) \cite{Goodfellow:2014:GAN:2969033.2969125} and Generative Query Network (GQN) \cite{DeepMind}. Figure~\ref{fig1} shows the overall structure of our method and compares it to the previously proposed GANs and GQN.

\subsection{Generative Adversarial Networks (GANs)} \label{GAN_intro}
    Generative Adversarial Networks (GANs) \cite{Goodfellow:2014:GAN:2969033.2969125} consist of two competing architectures referred as a generator (\textit{G}) and a discriminator (\textit{D}) (see Figure~\ref{fig1}(a)). The generator \textit{G} maps a given latent representation $z$ (possibly a vector with random values) into a novel image $x'=G(z)$ that is then passed to the discriminator network \textit{D}. The discriminator aims to determine if the given sample is produced by the generator, or if it is a real image taken from the training set. Denoting the real training samples as $x$, the conventional generator loss $\mathcal{L}_{G}^{GAN}$ and discriminator loss $\mathcal{L}_{D}^{GAN}$ are defined as:
  \begin{align}
\mathcal{L}_G^{GAN} &= -\mathbb{E}_{\bm{z} \thicksim P_{\bm{z}}}[\log{D(G(\bm{z}))}] \label{eqn:L_G} \\
\mathcal{L}_D^{GAN} &= -\mathbb{E}_{\bm{x} \thicksim P_{x}}[\mathrm{log} D(\bm{x})] - \mathbb{E}_{\bm{z} \thicksim P_{\bm{z}}}[\mathrm{log} (1 - D(G(\bm{z}))] \label{eqn:L_D}
\end{align}  
    Both networks are trained simultaneously in an alternating fashion. In the ideal case, the procedure guides the generator to produce images that are indistinguishable from the training image distribution. However, in practice the training procedure is challenging due to various problems such as mode collapses \cite{UnrollGAN}. 
     
\subsection{Generative Query Network (GQN)}\label{GQN_intro}
    Generative Query Network (GQN) \cite{DeepMind} is a deep generative model that learns the scene representation to perform novel view synthesis. Using an arbitrary number of observations, GQN can be trained to generate new views from the same environment. A GQN network includes a scene encoder(\textit{Enc}) and a decoder(\textit{Dec}) as can be seen in Figure~\ref{fig1}(b). \par
    First, the \textit{Enc} network tries to compress all of the 3D scene information into a latent representation $z$ from multiple pairs of input views $x$ and their camera poses $p$. The \textit{Enc} network processes each pair of input views and camera poses by using a feed-forward deep convolutional neural network. Each camera pose $p$ is represented by a 7 dimensional vector of $x_t$, $y_t$, $z_t$, $sin(yaw)$, $cos(yaw)$, $sin(pitch)$ and $cos(pitch)$ where $x_t$, $y_t$, $z_t$ are 3D translation, and $yaw$ and $pitch$ are parameters of the 3D rotation matrix. The latent scene information $z$ is the summation of all output pairs from the \textit{Enc} network. \par
    The goal of the GQN is to predict novel views by using the queried pose $p_q$. Therefore, $p_q$ and $z$ are input to the \textit{Dec} network to generate the new view $x'$. The \textit{Dec} network is a conditional latent variable model DRAW \cite{DRAW} which includes $M$ pairs of Inference and Generation recurrent sub network \cite{convolutional_DRAW}. In the language of GANs, the term of decoder \textit{Dec} and generator \textit{G} have many similarities since they both synthesize data from the latent vector $z$ (could be also random vector). Both \textit{Enc} and \textit{Dec} are trained jointly to minimize the evidence lower bound (ELBO) loss $\mathcal{L}_{GQN}$ :
\begin{equation}
\label{eqn:L_GQN}
\mathcal{L}_{GQN} = -\mathbb{E}_{(x,p),z}\bigg[
    -\ln\mathcal{N}(x_{gt}|x')+\sum_{n=1}^{M} \Big[\mathcal{N}(q_m,{\pi}_m)\Big]
\bigg]
\end{equation}
    The GQN training loss $\mathcal{L}_{GQN}$ is the expected value over the negative log-likelihood of the target image $x_{gt}$ given the target distribution regularized by the cumulative Kullback Leibler divergence between obtained posterior $q_m$ and prior $\pi_m$ distributions from $m^{th}$ generation step.
    
\section{Generative Adversarial Query Network}\label{GAQN}
The proposed Generative Adversarial Query Network (GAQN) builds on the GQN architecture by introducing two novel contributions. We will explain both of these in the following subsections. The corresponding experimental results are presented in Section \ref{Experiments}. Figure~\ref{fig1}(c) shows the overall architecture of our method.

\subsection{Adversarial loss}\label{GAQN_withLSGANloss}
As illustrated in Figure~\ref{fig1}(c), the proposed GAQN consists of three components: an encoder network \textit{Enc}, a decoder network \textit{Dec}, and a discriminator network \textit{D}. The GAQN utilities the same \textit{Enc} and \textit{Dec} architecture as standard GQN to generate a novel view $x'=Dec(Enc(x,p),p_q)$. However, inspired by the recent advancement of GANs, we include an additional discriminator network \textit{D} to distinguish between the generated fake images from the GQN and the ground truth view in the training data. In this way, the discriminator acts as a learn-able loss function that boosts the learning process of GQN. 

Numerous works \cite{WGAN,UnrollGAN,Loss-SensitiveGAN,ImprovedGAN} have shown that the training of the GAN may be unstable due to the vanishing gradients problem caused by the binary cross entropy loss as defined in Equation (\ref{eqn:L_G}) and (\ref{eqn:L_D}). In this paper, we avoid the above problem by adopting the least-square loss function from the previously proposed Least Squares Generative Adversarial Networks (LSGANs) \cite{LS_GAN_new}. The idea of LSGANs has been proved to be effective since it tries to pull the fake samples closer to the decision boundary of the least-square loss function. Based on the distance between the sampled data and the decision boundary, LSGANs manages to generate better gradients to update its generator. Furthermore, LSGANs also proves to exhibit less mode-seeking behaviour \cite{PR_book} which also stabilize the training process. Equation~(\ref{eqn:L_LSGAN_G}) and~(\ref{eqn:L_LSGAN_D}) provide the generator and discriminator loss of LSGANs that we use to train the GQN decoder and our proposed discriminator, respectively. \par

\begin{align}
\mathcal{L}_G^{LSGAN} &= \mathbb{E}_{\bm{z} \thicksim P_{\bm{z}}}\Big[(D(G(\bm{z}))-1)^2\Big] \label{eqn:L_LSGAN_G} \\
\mathcal{L}_D^{LSGAN} &= \mathbb{E}_{\bm{z} \thicksim P_{\bm{z}}}\Big[(D(G(\bm{z})))^2\Big] + \mathbb{E}_{\bm{x} \thicksim P_{\bm{x}}}\Big[(D(x')-1)^2\Big] \label{eqn:L_LSGAN_D}
\end{align}

Inspired by recent works~\cite{SAGAN,SpecGAN} on GANs, our proposed Discriminator $D$ network adopts the residual blocks architecture \cite{Resnet}. Instead of directly classifying the generated image as real or fake, we follow the patch-based discriminator \cite{PatchGAN} to restrict the attention to the structure in the local patches. Table~\ref{table_Discriminator_architecture} shows the design of our discriminator network.

\begin{figure}[ht]
    \centering
    \includegraphics[width=\textwidth]{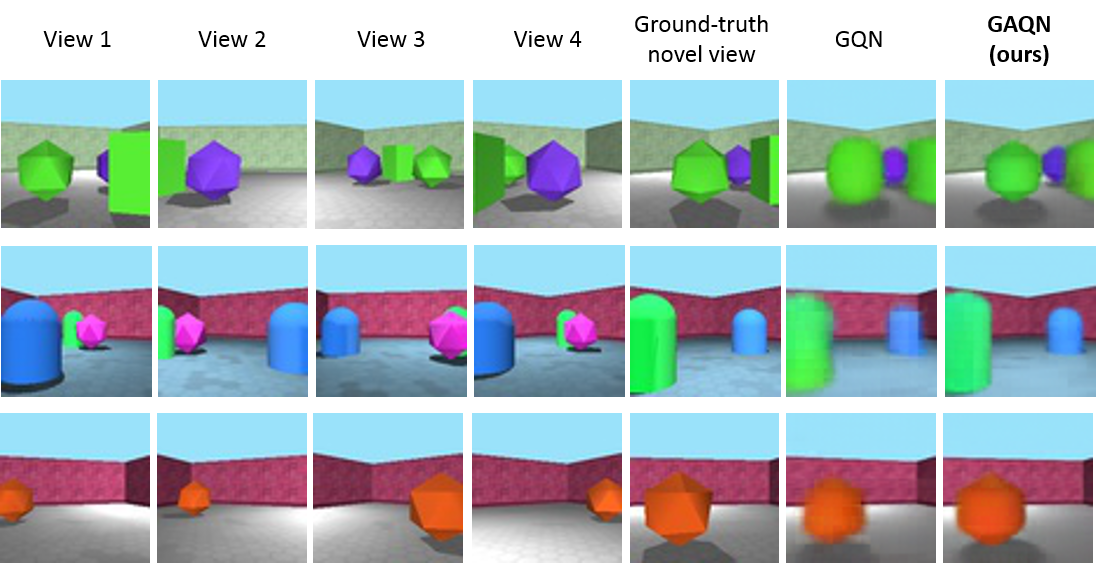}
    \caption{Comparison between generated novel view using the GQN and the proposed GAQN architectures. The first four columns depict the training views of the scene and the fifth column shows the ground truth image for the novel view. The last two columns illustrate the results obtained using GQN architecture and the proposed approach, respectively. Our method is able to obtain clearly sharper images compared to the plain GQN architecture.}
    \label{fig2}
\end{figure}

The original GQN includes per-pixel variance annealing technique \cite{DeepMind} that is aimed to focus the learning for the scene environment (wall, floor and sky), location and color of objects at the early stage of the training and enhance the low level details later. Even though the annealing strategy was shown to work, we argue that the adversarial loss is able to further speed up the learning and to generate sharper object details at the early stages of the training. Therefore, our GAQN  can achieve better visualized novel view earlier than GQN. 

Figure~\ref{fig2} illustrates the predicted novel view using the plain GQN architecture and the proposed GAQN architecture. Although the obtained GQN model successfully predicts the correct location and color of three objects in the 3D scene, their edges are blurry. Meanwhile our proposed method manages to produce sharper object edges, especially, to the middle green icosahedron. Further results and implementation details are provided in Section \ref{Experiments}. \par

\begin{table}[ht]
\centering
\caption{Architecture of our discriminator $D$ network using 3 residual blocks~\cite{Resnet}.}
\label{table_Discriminator_architecture}
\begin{tabular}{|c|c|c|}
\hline
{\textbf{Layers}} & \textbf{Input size} & \textbf{Output size} \\ \hline
conv\(2 \times 2\), stride = 2, channels = 32, ReLU           & \(64 \times 64 \times 3\)         & \(32 \times 32 \times 32\)         \\ \hline
ResBlock down 64                       & \(32 \times 32 \times 32\)        & \(16 \times 16 \times 64\)         \\ \hline
ResBlock down 128                      & \(16 \times 16 \times 64\)        & \(8 \times 8 \times 128\)          \\ \hline
ResBlock down 256                      & \(8 \times 8 \times 128\)         & \(4 \times 4 \times 256\)          \\ \hline
ResBlock down 512                      & \(4 \times 4 \times 256\)         & \(2 \times 2 \times 512\)          \\ \hline
conv\(1 \times 1\), stride = 1, channels = 1024, ReLU       & \(2 \times 2 \times 512\)         & \(2 \times 2 \times 1024\)         \\ \hline
\end{tabular}
\end{table}

\subsection{Feature-matching loss}\label{GAQN_FeatureMatching}
Inspired by the recent works \cite{CVAE_GAN,ImprovedGAN,ImprovingGANbyFM} on improving the stability of the GAN training, we add an extra feature matching loss to train the generator network. The main idea of this feature matching loss is to use the discriminator network as a feature extractor and guide the generator to generate data that matches the feature statistics of the real data. There are several approaches on exploiting the feature matching loss in training the generator.\par 
Specifically, the common GAN generator loss as shown in Equation~(\ref{eqn:L_G}) is being replaced by a mean feature matching loss. It has been argued that this mean feature matching loss helps preventing the gradient vanishing problem during the training. In our research, we have already adopted the least square loss to address the above problem (Section~\ref{GAQN_withLSGANloss}) but there is no guarantee that the problem is completely solved. Therefore, we train our GAQN generator network using a unified loss function $\mathcal{L}_{G}^{GAQN}$ as the combination of LSGANs generator loss $\mathcal{L}_{G}^{LSGAN}$ and mean feature matching loss $\mathcal{L}_{FM}$. Let $f_{D}()$ be the mean of the output feature maps from the 3$^{rd}$ layer (ResBlock down 128 in Table~\ref{table_Discriminator_architecture}) of the discriminator network, the mean feature matching loss is define as follow:
\begin{equation}
\label{eqn:L_FM}
\mathcal{L}_{FM} = ||\mathbb{E}_{\bm{x} \thicksim P_{\bm{x}}}f_{D}(x_{gt})-\mathbb{E}_{\bm{z} \thicksim P_{\bm{z}}}f_{D}(x')||_2^2
\end{equation}
In this paper, we jointly train GAQN \textit{Enc} and \textit{G} by using $\mathcal{L}_{EG}^{GAQN}$ which is the conventional GQN ELBO loss (Equation (\ref{eqn:L_GQN})). The \textit{D} parameters are being updated by the least squares loss (Equation (\ref{eqn:L_LSGAN_D})) from LSGANs, and we adopt the mean feature matching loss (Equation (\ref{eqn:L_FM})) to update \textit{G}. Finally, Equations (\ref{eqn:L_GAQN_G}), (\ref{eqn:L_GAQN_D}) and (\ref{eqn:L_GAQN_EG}) show all the loss functions we have used to train our GAQN model.\par

\begin{align}
\mathcal{L}_G^{GAQN} &= \mathcal{L}_G^{LSGAN} + \mathcal{L}_{FM} \label{eqn:L_GAQN_G} \\
\mathcal{L}_D^{GAQN} &= \mathcal{L}_D^{LSGAN} \label{eqn:L_GAQN_D} \\
\mathcal{L}_{EG}^{GAQN} &= \mathcal{L}_{GQN} \label{eqn:L_GAQN_EG}
\end{align}

\section{Experiments} \label{Experiments}

\subsection{Experimental settings} \label{exp_settings}
We evaluate our method using the \textit{rooms ring camera} dataset that was provided and used by Eslami et al. in \cite{DeepMind}. The dataset contains various rendered 3D square rooms that contain random objects of various shapes, colors and locations. Moreover, the scene textures, walls and lights are also randomly generated. Therefore, the task of predicting the novel views on this data-set is a relatively challenging problem. 

The original GQN construction proposed in \cite{DeepMind} consumes a large amount of computing and memory resources. Due to computational restrictions, we demonstrate the advantages of the proposed architectural changes using a smaller version of the basic GQN network and a subset of the training data. Originally, training a 12 generative-layers GQN model with batch size of 36 requires 4 NVIDIA K80 GPUs  as shown in \cite{DeepMind}. In this paper, we use a single NVIDIA Tesla P100 GPU to train our GAQN model which has 8 generative layers and batch size is 20. As far as we experimented, this change would not affect the quality of results. Despite of the smaller network size, our method is able to produce results (see figures~\ref{fig2} and~\ref{fig_final}) that are very close to those presented in the original work \cite{DeepMind} that uses clearly larger network and training set. This further, emphasized the benefits of the proposed architectural modifications. Moreover, we only use the first halve of the original GQN's training and testing data for faster training procedure. 

 We implement our model on PyTorch~\cite{pytorch} and our GAQN model is trained end-to-end using ADAM optimization \cite{ADAM} with hyper-parameter $\beta_1=0.9$ and $\beta_2=0.999$. The generator network is trained using the learning rate of $10^{-4}$ and the discriminator is trained using a learning rate of $4 \times 10^{-4}$. Recent work by Heusel et al.~\cite{TTUR} shows that if we update the generator slower than the discriminator then it helps to reach the convergence easier. 

\subsection{Results}

\begin{table}[ht]
\centering
\caption{Comparison of training, testing loss, KL testing loss and SSIM testing score between our GAQN model (GQN + LSGANs + FM), original GQN and a variant (training GQN with LSGANs loss).}
\label{table_Compare_methods}
\begin{tabular}{c|c|c|c|}
\cline{2-4}
                                               & GQN   & GQN + LSGANs & \textbf{\begin{tabular}[c]{@{}c@{}}GAQN\\ (ours)\end{tabular}}  \\ \hline
\multicolumn{1}{|c|}{\textbf{Training loss}}   & 7012  & 6988      & \textbf{6951}  \\ \hline
\multicolumn{1}{|c|}{\textbf{Testing loss}}    & 7003  & 6974      & \textbf{6957}  \\ \hline
\multicolumn{1}{|c|}{\textbf{KL testing loss}} & 17.59 & 19.36     & \textbf{22.91} \\ \hline
\multicolumn{1}{|c|}{\textbf{SSIM}} & 0.742 & 0.815     & \textbf{0.865} \\ \hline
\end{tabular}
\end{table}

\begin{figure}[!ht]
    \centering
    \includegraphics[width=\textwidth]{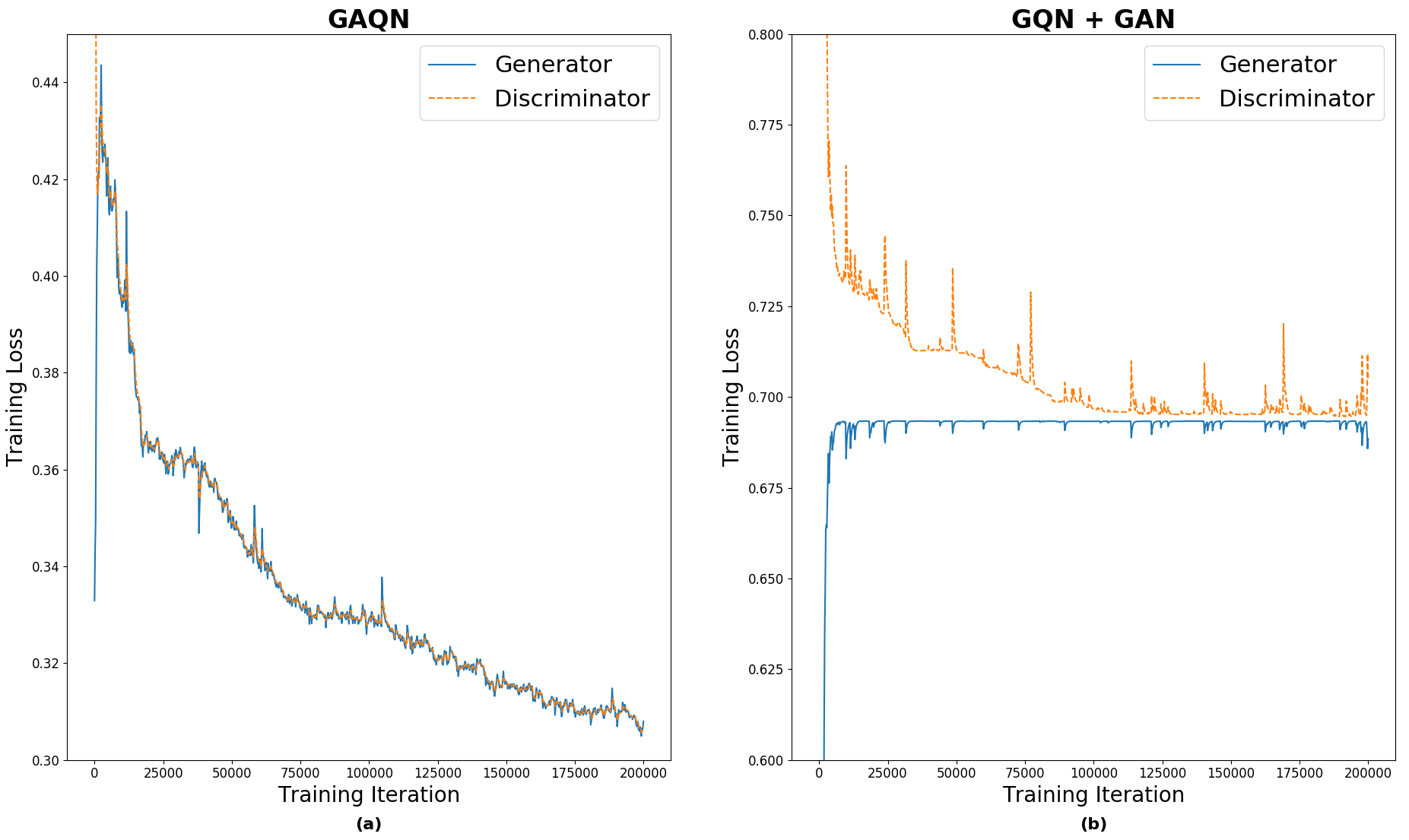}
    \caption{Comparison of generator and discriminator training loss between our proposed GAQN (a) and GQN + GAN (b). Both generator and discriminator of GQN+GAN are suffering from mode collapsing and vanishing gradients. Using the least-square loss and the mean feature loss, our GAQN achieves a stable learning process.}
    \label{fig_compare}
\end{figure}

\begin{figure}[!ht]
    \centering
    \includegraphics[width=\textwidth]{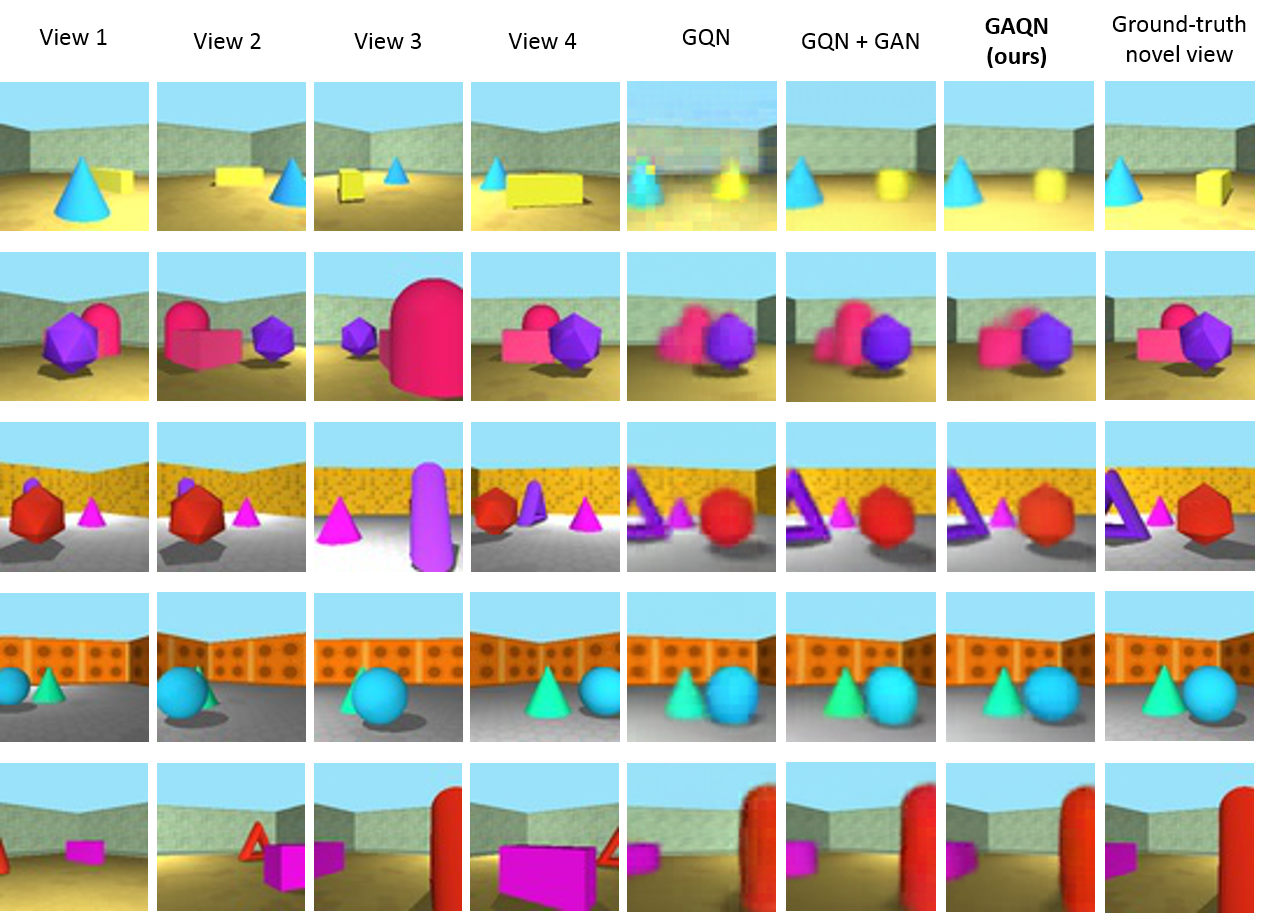}
    \caption{Generated novel view comparison between our proposed GAQN and variants}
    \label{fig_final}
\end{figure}

Our approach relies on two new components, namely, an adversarial training pipeline using the least-square GANs loss and the discriminator feature matching loss functions to enhance the previously proposed GQN model. In this section, we will experimentally investigate the impact of the both components. The results corresponding to only GQN and adversarial loss are denoted as (GQN + LSGANs), whereas the results with the proposed full model are denoted as GAQN. We train the above methods using the same hyper-parameters (reported in Section~\ref{exp_settings}) and record their training and testing loss for evaluation purposes. We also use the structure similarity index (SSIM) to assess on the quality of the target image and the predicted novel on a held-out test set.

Table~\ref{table_Compare_methods} contains the obtained results. The proposed full GAQN method has clearly the smallest training and testing loss and the largest Kullback–Leibler (KL) testing loss. The KL loss represents the distance between the estimated and true distribution produced by GQN's generation network and this KL loss is the second term in the ELBO loss (Equation (\ref{eqn:L_GQN})). If the model manages to produce high KL loss then the predicted novel view tends to be close to the ground-truth data. It is also evident that the adversarial loss is able to improve the results compared to the plain GQN architecture. However, the largest gain is obtained by combining both of the proposed contributions. 

In Figure~\ref{fig_final}, we show qualitative examples of the generated novel views produced by different versions of the proposed GAQN method and the baseline. Although the baseline manages to correctly generate the colors and positions of objects in most cases, theirs edges are blurry. Based on the SSIM, our GAQN model achieves highest score and produce sharper edges on predicted images. \par

Finally, we illustrate how the least squares loss of LSGANs and discriminator feature matching loss affect the training procedure of the method. In this experiment, we compare the generator and discriminator training loss of our proposed GAQN and GQN + GAN. As can be seen in Figure~\ref{fig_compare} (b), the training procedure of GQN + GAN is highly unstable due to mode collapsing and diminishing gradients. Our GAQN model eliminates both problems by using the least-squares loss and the discriminator mean feature loss as shown in Figure~\ref{fig_compare} (a).

\section{Conclusion} \label{Conclusion}
We have introduced an novel adversarial training pipeline to improve the previously proposed GQN network architecture. Our experimental results demonstrate that training an additional discriminator network encourages the GQN model to predict more accurate novel views. Moreover, using the combination of least square loss and the feature matching loss helps stabilizing both generator and discriminator training process. In our future work, we will explore how to generate novel views in the bigger scale of indoor scenes where there are more objects and different lighting conditions.
\bibliographystyle{splncs04}
\bibliography{refs} 
\end{document}